\documentclass[sigconf]{acmart}
\usepackage[ruled]{algorithm2e}
\usepackage{multirow}

\AtBeginDocument{
  }

\copyrightyear{2026}
\acmYear{2026}
\setcopyright{cc}
\setcctype{by}
\acmConference[XXX]{XXX}{XXX}{XXX, XXX}
\acmBooktitle{XXX}
\acmPrice{}
\acmDOI{XXX}
\acmISBN{XXX}

\begin{document}

\title[LLM-Upgraded Graph Reinforcement Learning for Carbon-Aware Job Scheduling]{LLM-Upgraded Graph Reinforcement Learning for Carbon-Aware Job Scheduling in Smart Manufacturing}

\makeatletter
\def\@affiliationfont{\fontsize{9.5pt}{11pt}\selectfont}
\makeatother

\author{Zhiying Yang}
\affiliation{
  \institution{Singapore Institute of Technology}
  \country{Singapore}}
\email{zhiying.yang@singaporetech.edu.sg}

\author{Fang Liu}
\affiliation{
  \institution{Singapore University of Social Sciences}
  \country{Singapore}}
\email{liufang@suss.edu.sg}

\author{Wei Zhang}
\authornote{Corresponding author.}
\affiliation{
  \institution{Singapore Institute of Technology}
  \country{Singapore}}
\email{wei.zhang@singaporetech.edu.sg}

\author{Xin Lou}
\affiliation{
  \institution{Singapore Institute of Technology}
  \country{Singapore}}
\email{lou.xin@singaporetech.edu.sg}

\author{Malcolm Yoke Hean Low }
\affiliation{
  \institution{Singapore Institute of Technology}
  \country{Singapore}}
\email{malcolm.low@singaporetech.edu.sg}

\author{Boon Ping Gan}
\affiliation{
  \institution{D-SIMLAB Technologies}
  \country{Singapore}}
\email{boonping@d-simlab.com}

\renewcommand{\shortauthors}{Z. Yang et al.}

\begin{abstract}
This paper presents \textsc{Luca}, a \underline{l}arge language model (LLM)-\underline{u}pgraded graph reinforcement learning framework for \underline{c}arbon-\underline{a}ware flexible job shop scheduling. \textsc{Luca} addresses the challenges of dynamic and sustainable scheduling in smart manufacturing systems by integrating a graph neural network and an LLM, guided by a carefully designed in-house prompting strategy, to produce a fused embedding that captures both structural characteristics and contextual semantics of the latest scheduling state. This expressive embedding is then processed by a deep reinforcement learning policy network, which generates real-time scheduling decisions optimized for both makespan and carbon emission objectives. To support sustainability goals, \textsc{Luca} incorporates a dual-objective reward function that encourages both energy efficiency and scheduling timeliness. Experimental results on both synthetic and public datasets demonstrate that \textsc{Luca} consistently outperforms comparison algorithms. For instance, on the synthetic dataset, it achieves an average of 4.1\% and up to 12.2\% lower makespan compared to the best-performing comparison algorithm while maintaining the same emission level. On public datasets, additional gains are observed for both makespan and emission. These results demonstrate that \textsc{Luca} is effective and practical for carbon-aware scheduling in smart manufacturing.
\end{abstract}



\maketitle

\section{Introduction}
Manufacturing is a critical economic pillar. It accounts for 18\% of the gross domestic product (GDP) of Singapore and over one quarter of China's GDP \cite{worldbank_sg_manu}. The percentages are also significant in many other nations. Within this domain, job shop scheduling (JSP) directly impacts productivity, resource utilization, and system responsiveness to demand. Ineffective scheduling can lead to 5$-$20\% productivity loss based on \cite{source_downtime_loss}. The focus of traditional scheduling solutions has been speed-driven, e.g., makespan and throughput, while modern manufacturing industry face demands broader than speed, and new considerations include real-time decision-making, sustainability, etc. As industry evolves, so does the need for adaptive and multi-objective scheduling that aligns with efficiency and environmental targets.

A key challenge in modern manufacturing lies in scheduling for complex systems. Flexible JSP (FJSP) captures the operational complexity where each job consists of a number of operations each can be assigned to one or multiple machines. In carbon-aware FJSP, machines are also characterized by different carbon emission profiles and scheduling is treated as a multi-objective problem involving both makespan minimization and emission reduction. While real-world manufacturing complexity has been captured by the carbon-aware FJSP, solving the problem requires more than basic solutions like numerical optimization. Emerging needs include adaptive strategies that respond to dynamic manufacturing contexts and representations that reflect scheduling priorities, to achieve optimal scheduling.

Existing works for solving JSP including FJSP range from classic heuristics to machine learning (ML) based methods. Early efforts include heuristics such as \textsc{Fifo} \cite{chen2013flexible}, \textsc{Mwkr} \cite{brandimarte1993routing} and \textsc{Spt} \cite{montazeri1990analysis}, which offer simplicity but lack adaptability in complex system dynamics and multi-object scheduling contexts. Recent advancements are mostly ML-based, e.g., carbon-aware evolutionary algorithms \cite{liu2017hybrid,piroozfard2018minimizing} and carbon-aware energy scheduling \cite{zhao2025diffusion}. Among them, \textsc{Drl} \cite{song2022flexible} is one of the state-of-the-art (SOTA) methods. It leverages a hybrid model based on graph neural network (GNN) and reinforcement learning (RL). Despite its competitive performance in makespan minimization, \textsc{Drl} can only handle one objective and is carbon agnostic. Also it processes numerical features only, similar to \cite{zhao2024actor,zhang2024deep,wang2022cea}, and overlooks the potential of symbolic-level model upgrade to well capture evolving manufacturing contexts.

In this paper, our idea is to use RL to generate scheduling policies, a direction shown to be effective in solving FJSPs. One challenge, however, lies in designing an effective representation of the scheduling state. The representation shall be expressive and responsive to real-time system dynamics, the description of which can be numerical- and text-based. To address this, we explore the integration of recent advances in ML, specifically GNN and large language models (LLMs) \cite{zhang2024role}. We expect the fusion to enable a semantically rich and context-aware state representation, to optimize the scheduling process within the RL framework. There are early attempts of using LLMs in FJSP, e.g., \cite{abgaryan2024llms}. However, the reported performance still lags behind the SOTAs like \textsc{Drl}, highlighting the technical challenges of using new technologies effectively.

Specifically, we propose \textsc{Luca}, an \underline{L}LM-\underline{u}pgraded graph deep RL for \underline{c}arbon-\underline{a}ware FJSP. In \textsc{Luca}, a job scheduling instance is first encoded into a graph, capturing jobs and operations, precedence dependencies, and machines and carbon factors. The graph changes over time to reflect the latest scheduling and execution state of the instance. A GNN processes this graph to extract a structural and numerical embedding. The graph is also described in the text format and the text is used as the prompt of an LLM to generate a contextual embedding, which is expected to be expressive based on the LLM's pattern recognition and reasoning capabilities. The embeddings can be complementary and are passed to a gate fusion module to generate a unified embedding or representation of the latest scheduling state. Besides, \textsc{Luca} incorporates a new reward function based on both makespan and emission to guide its policy training, and the trained policy optimizes both objectives together. Finally, \textsc{Luca} determines a scheduling action by processing the fused embedding with its policy network. 

We have conducted a comprehensive experimental study. \textsc{Luca}'s performance is compared with several comparison algorithms based on both synthetic and public datasets. On the synthetic dataset, \textsc{Luca} outperforms the second-best performing algorithm by 4.1\% on average and up to 12.2\% for makespan by maintaining the same level of emission. On public datasets, \textsc{Luca}'s advantage is more significant, with 6.6\% and 3.4\% improvement for makespan and emission, respectively. The experiment results showcase \textsc{Luca}’s potential as a practical and competitive job scheduling and decision-making solution, which contributes toward efficient and sustainable manufacturing systems.
    
The remainder of this paper is organized as follows. We first introduce carbon-aware FJSP in Section \ref{sec:problem}. Then, we introduce the system architecture and methodology in Section \ref{sec:method}. Section \ref{sec:experiment} presents the experimental study and discussions. Finally, Section \ref{sec:conclusion} concludes the paper and suggests future works.

\section{Problem Formulation: Carbon-Aware FJSP}
\label{sec:problem}
A clear and precise problem formulation is essential for algorithm design and experimental study and we first formulate carbon-aware FJSP addressed in this paper. Unlike the traditional JSP, where each operation must be assigned to a specific machine, carbon-aware FJSP introduces machine flexibility and each operation can be processed by one of several eligible machines with different efficiency and environmental impact. Such flexibility better reflects real-world manufacturing systems such as smart factories, although it also introduces algorithm design complexity.

In FJSP, an instance of size $n \times m$ consists of $n$ jobs and $m$ machines. We denote the sets of jobs and machines as $\mathcal{J}$ and $\mathcal{M}$, respectively. Each job $J_i \in \mathcal{J}$ comprises a sequence of $k_i$ operations $\mathcal{O}_i = (O^i_1,\ldots,O^i_{k_i})$, subject to precedence constraints, e.g., $O^i_{k}$ can only be started after the full completion of $O^i_1,\ldots,O^i_{k-1}$. Each operation $O^i_k$ can be processed by any one of the machines in a compatible subset $\mathcal{M}^{i,k} \subseteq \mathcal{M}$. Once a machine $M_j \in \mathcal{M}^{i,k}$ is selected for an operation, the associated processing time is denoted by $p^{i,k}_j$. All machines are non-preemptive and can handle only one operation at a time. Besides time, we further incorporate carbon factors into our formulation to address the growing need for sustainability in smart manufacturing. Specifically, each machine $M_j$ is associated with a carbon emission rate $e_j$, which can be different for different machines. 

Given an instance, a \textit{schedule} specifics an execution plan of each operation $O^i_k$, including machine $M_j$ on which $O^i_k$ is processed, the start time $T^{\text{start}}_{i,k}$, and completion time $T^{\text{end}}_{i,k} = T^{\text{start}}_{i,k} + p^{i,k}_j$. A job is considered as complete when all its associated operations have been completed. The \textit{makespan} of the schedule is defined as the time required to complete all operations of the instance as,
\begin{equation} \label{eq:makespan}
    T^{\max} = \max_{J_i \in \mathcal{J}} \{ T^{\text{end}}_{i} \},
\end{equation}
where $T^{\text{end}}_{i}$ is the time of completing the last operation of job $J_i$ with operations $O^i_1,\ldots,O^i_{k_i}$, i.e., $T^{\text{end}}_{i} = T^{\text{end}}_{i,k_i}$. Accordingly, the total carbon emission of a schedule is computed by summing the emissions generated by the machines for processing all operations across all jobs. Specifically, the \textit{emission} for executing operation $O^i_k$, which is assigned to machine $M_j$, is calculated as $p^{i,k}_j e_j$. The total emission is calculated as,
\begin{equation} \label{eq:emission}
    E^{\text{total}} = \sum_{J_i \in \mathcal{J}}\Big(\sum_{k_i} p^{i,k}_j e_j \Big).
\end{equation}

The overall objective of solving the carbon-aware FJSP is to find a schedule for an instance to minimize both makespan and carbon emission. The two objectives can be conflicting sometimes and complementary in certain conditions. 

\section{\textsc{Luca}: Methodology}
\label{sec:method}
Having formulated the carbon-aware FJSP, we now present our proposed \textsc{Luca}. We begin with an overview of \textsc{Luca}’s system architecture to highlight its key modules and workflow. Then, we present the technical details of each module.

\begin{figure*}[t]
    \centering
    \includegraphics[width=0.8\linewidth]{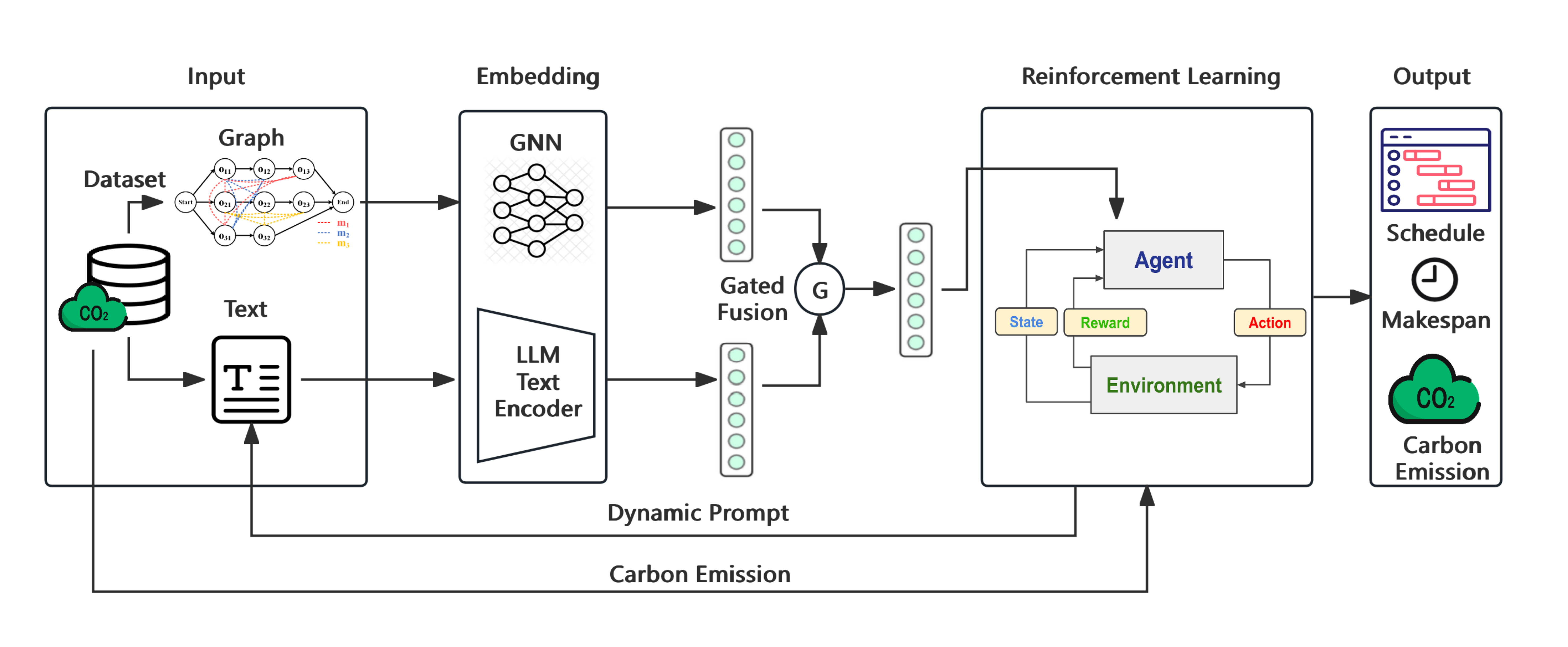}
    \caption{System architecture of \textsc{Luca} for carbon-aware FJSP. The framework leverages GNN and LLM to produce a structural and contextual embedding of a scheduling state, and RL to learn policies that jointly optimize makespan and carbon emissions.}
    \label{fig:system}
\end{figure*}

\subsection{System Architecture Overview} \label{sec:architecture}
The overall system architecture of \textsc{Luca} is illustrated in Figure \ref{fig:system}. \textsc{Luca} in general is RL-based. RL is well-suited for solving FJSP. It models the scheduling process as a sequential decision-making problem and uses an agent to learn adaptive policies that respond to time-varying scheduling states. A typical RL framework consists of several components, including states, actions, rewards, a policy, and a critic. We briefly describe them within the RL-based \textsc{Luca}.

In \textsc{Luca}, a state represents the real-time dynamics of the FJSP scheduling and execution. It captures the progress of each operation of each job, machine availability, and other factors such as carbon emission. \textsc{Luca} models such a state with two parallel encoding schemes. First, we represent an instance as a graph where nodes correspond to operations and edges represent precedence constraints. The graph evolves in real-time, e.g., completion status of each operation and associated carbon emission. We apply GNN to this evolving graph to encode the state and produce an embedding which contains the instance's operation features and scheduling context, e.g., precedence constraints. In parallel, \textsc{Luca} constructs a textual description of the current state, to emphasize semantic and contextual information that may not be explicitly encoded in the graph. A pre-trained LLM is used to encode this text-based description to generate an embedding that complements the GNN-generated embedding. Then, we fuse the two embeddings into a unified state using a gated fusion mechanism. 

A policy defines how the RL agent maps a given state to an action. In \textsc{Luca}, the fused embedding serves as the input to the policy network for making scheduling decisions. The agent is trained using the proximal policy optimization (PPO) algorithm to optimize the policy network with respect to minimizing makespan and emission, based on which the rewards are computed. \textsc{Luca} also periodically updates the prompt for the LLM-based embedding based on historical rewards, so that the policy benefits from more informed and adaptive representations over time. Overall, through this learning process, \textsc{Luca} is expected to produce a high-quality schedule that optimizes both operational efficiency and sustainability within a unified decision-making framework. 

\subsection{LLM-based Embedding}\label{sec:LLM}
\textsc{Luca} relies on LLM to capture an instance’s real-time semantic context through text-based state description and encodes operation-level descriptions into contextual embeddings. We now detail the process of LLM-based embedding and embedding integration into job scheduling. 

\subsubsection{State Prompt}
At each time slot $t$ for executing an instance, there are a set of completed operations and a set of uncompleted operations. Two tasks of \textsc{Luca} at the time slot $t$ are selecting an operation to execute at time slot $t+1$ and assigning the operation to a machine. We plan to describe the current scheduling state with text and use an LLM-based text encoder to convert the text into numerical representations. Our focus is the set of uncompleted operations. For each uncompleted operation $O^i_k$ in job $J_i$, we define $x_{i,k}^{\text{prompt}}$ as the state prompt of the operation and the prompt includes the instance and execution details of the operation. We create a prompt template to standardize and generate the text-based state description of an operation and one example is shown below. 
\begin{quote}\small
\centering 
\begin{ttfamily}
\{Job 0, Op 0, 5 ops left; est\_start=0.0, dur=7.4; machines=0:7.0\,$|$\,1:6.0\,$|$\,2:9.0\,$|$\,3:9.0\,$|$\,4:6.0\},
\end{ttfamily}
\end{quote}
where the prompt encodes the job identifier, operation identifier, remaining number of operations, the earliest start time, expected processing duration, along with the list of eligible machines, each specified by its identifier and corresponding processing time. The emission rate of each machine is implicitly represented in the prompt through its association with the machine identifier.

\subsubsection{Feedback Prompt}
In addition to the state prompt, we introduce a feedback prompt for each operation to explicitly include the operation's impact on scheduling objectives, i.e., makespan and emission. While state prompt already encodes the impact partially and implicitly, it may lack explicit information about an operation's contribution to these key objectives. We find that explicitly providing this feedback improves \textsc{Luca}'s learning. This aligns with how LLMs process input, especially for a lightweight LLM with limited reasoning capacity due to realistic industry constraints, that explicit and structured cues are more effective than implicit patterns.

During the training of \textsc{Luca}, each instance, similarly each operation of the instance, is used and reused in different training iterations to refine the model iteratively. Here, we save an operation's impact log to record the operation's makespan and emission in each iteration. An example of such a log is,
\begin{quote}\small
\centering
\begin{ttfamily}
\{op\_id: 29, makespan\_impact: $6.0$, emission\_impact: $95.0$\},
\end{ttfamily}
\end{quote}
where makespan and emission can be calculated based on the machine that the operation is assigned to. Let $\Delta^{\text{ms}}_{i,k,l}$ and $\Delta^{\text{ce}}_{i,k,l}$ represent the logged makespan and emission, respectively, for operation $O^i_k$ at iteration $l$. After certain number of iterations, e.g., $n_l$, we calculate the average impact of each operation $O^i_k$ as,
\begin{equation} \label{eq:ave_makespan}
    \delta_{i,k}^{\text{ms}}=\frac{1}{n_l}\sum_{l=1}^{n_l}\!\Delta^{\text{ms}}_{i,k,l}, \quad
    \delta_{i,k}^{\text{ce}}=\frac{1}{n_l}\sum_{l=1}^{n_l}\!\Delta^{\text{ce}}_{i,k,l}.
\end{equation}

We also introduce two pre-defined thresholds $\tau_{\text{ms}}$ and $\tau_{\text{ce}}$, for makespan and emission, respectively. And we generate hints about the operation's impact as,
\begin{equation}
\begin{cases}
    \delta_{i,k}^{\text{ms}} > \tau_{\text{ms}} \Rightarrow \texttt{``Hint: High Makespan Impact''}\\
    \delta_{i,k}^{\text{ce}}>\tau_{\text{ce}} \Rightarrow \texttt{``Hint: High Emission Impact''}
\end{cases},
\end{equation}
where the operation $O^i_k$ is labeled for its high impact on makespan (emission) if the average makespan (emission) impact exceeds the makespan (emission) threshold. We expand the state prompt with such impact hints from the feedback prompt. Note that it is possible to include both hints for an operation, which may have a big impact on both makespan and emission.

\subsubsection{LLM}
We adopt a lightweight pre-trained Sentence Transformer in this study to demonstrate the feasibility of \textsc{Luca} in the scenarios with limited computing resources. Note that \textsc{Luca} is not restricted to a specific LLM and other LLMs can also be adopted with minimal configuration effort. Here, we concatenate the prompts of all operations into a single input sequence containing both state and feedback prompts of the operations at time $t$. This unified prompt is processed by the LLM to produce a fixed-length, e.g., 128, embedding, denoted as $\mathbf{z}_t^{\text{LLM}} = \texttt{LLM}(X^{\text{prompt}}_t)$, where $X^{\text{prompt}}_t$ is the concatenated unified prompt input. $\mathbf{z}_t^{\text{LLM}}$ encodes the latest scheduling state at time $t$ and it then serves as one of the key inputs to the policy network. 

\subsection{GNN-based Embedding}
LLM-based embeddings offer contextual understanding through text prompts, but they do not explicitly model structural dependencies such as operation precedence. Here, \textsc{Luca} also employs a GNN to encode the job graph, which is constructed from the real-time scheduling state including jobs, operations, machines, precedence dependencies, and execution status. We expect this GNN-based embedding to capture scheduling constraints and operation-machine relationships that align closely with the underlying structure of carbon-aware FJSP. GNN-based embedding has been studied in previous works \cite{song2022flexible}, so we do not duplicate the effort and follow existing workflows. Let $G_t$ be the graph structure of the instance at time $t$, we apply GNN to this graph and generate GNN-based embedding $\mathbf{z}_t^{\text{GNN}}=\texttt{GNN}(G_t)$. 

\subsection{Fusion of Embeddings}
We make full use of the information encoded in both job graphs and text descriptions and we have both LLM- and GNN-based embeddings $\mathbf{z}_t^{\text{LLM}}$ and $\mathbf{z}_t^{\text{GNN}}$ at time $t$. \textsc{Luca} combines both embeddings and adopt a gated fusion module for optimal integration, where unified embedding $\mathbf{h}_t$ is produced for downstream scheduling decisions. The embeddings may not be equally important and the module is expected not to over-commit to either embedding. Here, we design the module to compute the fused embedding as,
\begin{equation}
\begin{split}
&\mathbf{h}_t = g_t \cdot \texttt{proj}\big(\mathbf{z}_t^{\text{LLM}}\big) 
+ (1 - g_t) \cdot \mathbf{z}_t^{\text{GNN}}, \\
&g_t = \sigma\bigg(
W \Big[
\texttt{proj}(\mathbf{z}_t^{\text{LLM}}) \mathbin\Vert \mathbf{z}_t^{\text{GNN}}
\Big]
\bigg),
\end{split}
\end{equation}
where $g_t \in (0, 1)$ is a learnable gating value to control the importance of LLM- and GNN-based embeddings. A learnable projection layer $\texttt{proj}(\cdot)$ first maps the LLM embedding $\mathbf{z}_t^{\text{LLM}}$ to the same dimensional space as $\mathbf{z}_t^{\text{GNN}}$ to ensure compatibility. $\sigma(\cdot)$ denotes the \texttt{sigmoid} activation function, and $W$ is a learnable weight matrix that produces the gating value $g_t$ based on the concatenated embeddings $\texttt{proj}(\mathbf{z}_t^{\text{LLM}})\mathbin\Vert \mathbf{z}_t^{\text{GNN}}$ with concatenation operator $\mathbin\Vert$. Overall, the fusion module adapts in real-time to decide how much to rely on LLM-derived information versus GNN-based features. \textsc{Luca} benefits from such optimized and complementary information for improved scheduling.

\subsection{Carbon-Aware Policy}\label{sec:RL}
We aim to achieve effective and generalizable learning and we design and present \textsc{Luca}'s carbon-aware policy. We first focus on the policy formulation and then introduce a held-out validation mechanism to promote the policy generalization.

\subsubsection{Policy Formulation and Rewards}
We model the carbon-aware FJSP as a sequential decision-making process. At each time $t$, the agent of \textsc{Luca} observes the current scheduling state $s_t$, represented by embedding $\mathbf{h}_t$, and selects an action $a_t = (O^i_k, M_j)$, which means assigning operation $O^i_k$ to machine $M_j$. The environment transitions to the next state $s_{t+1}$ and returns a reward $R_t$ that quantifies the scheduling performance for makespan and emission. \textsc{Luca} uses PPO to learn a policy for selecting operation–machine pairs optimally based on the returned reward. The reward shall reflect the scheduling performance on both makespan and emission and we formulate the reward $R_t$ at time $t$ as,
\begin{equation} \label{eq:reward}
R_t = (1 - \lambda) \cdot \texttt{norm}(R_t^{\text{ms}}) +\lambda \cdot \texttt{norm}(R_t^{\text{ce}}), \quad \lambda \in [0,1],
\end{equation}
where $\lambda$ is a parameter to control the importance between makespan and emission, e.g., $\lambda = 0.5$ means that makespan and emission are equally important. $R_t^{\text{ms}}$ and $R_t^{\text{ce}}$ are the makespan and emission rewards, respectively. Specifically, $R^{\text{ms}}_t = \sum_{v=0}^{v'} \gamma^v r^{\text{ms}}_{t+v}$, where $\gamma$ is a discount factor between 0 and 1, $r^{\text{ms}}_t$ is the immediate makespan reward at time step $t$, and $v$ reflects the horizon of the reward calculation. The emission reward $R^{\text{ce}}_t$ follows the same calculation scheme. Note that the two rewards can be significantly different, e.g., by orders of magnitude. So, we normalize the rewards before applying the importance control parameter $\lambda$; specifically, \texttt{$z$-score} normalization is adopted. With the calculated composite reward $R_t$, we compute the advantage estimate as $R_t - V_\phi(s_t)$, where $V_\phi(s_t)$ is the value at state $s_t$ approximated by the critic network with parameters $\phi$. The advantage serves as a key component in the PPO loss function.

Let $\pi_\theta(\cdot)$ be the policy which is modeled by a neural network with parameters $\theta$. $\pi_\theta(a_t | s_t)$ means the probability of choosing action $a_t$ for operation-machine assignment given state $s_t$, under policy parameters $\theta$. To model this probability, we define a scoring function $P(a_t, s_t)$ that evaluates the quality of the pair of candidate action $a_t$ and current state $s_t$. The policy is then computed using a \texttt{softmax} distribution over all feasible actions $\mathcal{A}_t$ at time $t$ as,
\begin{equation} \label{eq:policy}
\pi_\theta(a_t | s_t) = \frac{\exp\big(P(a_t, s_t)\big)}{\sum_{a'_t \in \mathcal{A}_t} \exp\big(P(a'_t, s_t)\big)},
\end{equation}
where $P(a_t, s_t)$ is implemented as a neural network and the denominator ensures that the probabilities over all possible pairs sum to 1. During training, the parameters $\theta$ are updated to maximize the expected reward. The policy is expected to produce actions that contribute to the optimization of both objectives.

\subsubsection{Validation for Generalization}
The policy shall not be specialized to the training instances only, and we aim to promote policy generalization by maintaining a held-out replay buffer. This buffer, denoted as $\mathcal{I}_{\text{val}}$, contains unseen scheduling instances that are not used in the above policy training and serve solely for validation. We maintain model checkpoints at fixed intervals $l_{\text{check}}$, and each interval includes several iterations of training. At the end of each interval, we evaluate the latest policy $\pi_\theta$ on $\mathcal{I}_{\text{val}}$ by computing the average rewards for makespan and carbon emission, denoted as $\bar{R}^{\text{ms}}$ and $\bar{R}^{\text{ce}}$, respectively. These rewards are normalized and then summed according to Eq. (\ref{eq:reward}). Such an aggregated reward estimates the generalization performance of the policy for unseen scheduling instances. If the validation performance degrades compared to the beginning of the interval, we revert the policy to the beginning of the checkpoint. This strategy prevents policy degradation effectively during training by ensuring that only performance-improving updates are retained.

\begin{algorithm}[t] 
\caption{Training Procedure of \textsc{Luca}}
\KwIn{LLM encoder, GNN encoder, policy network $\pi_\theta$ and critic network $V_\phi$, thresholds $\tau_{\text{ms}}, \tau_{\text{ce}}$;} 

Sample a batch of $\mathcal{B}$ FJSP instances\;
\For{$iter = 1$ to $iter^{\max}$}{
    \For{each $b$ in $\mathcal{B}$}{
        Initialize LLM embedding $\mathbf{z}_0^{\text{LLM}}$ and GNN feature embedding $\mathbf{z}_0^{\text{GNN}}$\;
        Compute fused embedding $\mathbf{h}_0$ to represent initial state $s_0$\;
        \While{termination condition is not met}{
            Sample $a_t \sim \pi_\theta(\cdot|s_t)$;
            Compute reward $R_t$\;
            Compute the impact for the processed operation $O^i_k$\;

            \If{$iter \bmod n_l = 0$}{
                \If{$\delta_{i,k}^{\text{ms}} >  \tau_{\text{ms}}$}{add ``High Makespan Impact'' to $x^{\text{prompt}}_{i,k}$\;}
               
                \If{$\delta_{i,k}^{\text{ce}} >  \tau_{\text{ce}}$}{add: ``High Emission Impact'' to $x^{\text{prompt}}_{i,k}$\;} 
                }
                Update $X^{\text{prompt}}_{t+1}$ and encode LLM embedding $\mathbf{z}_{t+1}^{\text{LLM}}$\;
                Extract GNN feature embedding $\mathbf{z}_{t+1}^{\text{GNN}}$\;
                Compute fused embedding $\mathbf{h}_{t+1}$ to represent state $s_{t+1}$\;
            }
        }
        Compute advantages estimate $\Delta^R_t$, and update $\theta$ and $\phi$ via PPO\; 
        \If{$iter \bmod l_{\text{check}} == 0$}{validate the policy\;}  
        \If{$iter \bmod l_{\text{batch}} == 0$}{resample $\mathcal{B}$ FJSP instances\;}  
    }
\KwOut {trained policy $\pi_\theta$}
\label{alg:Luca}
\end{algorithm}

\subsection{Training Procedure}
The training procedure of \textsc{Luca} is presented in Algorithm \ref{alg:Luca}. Each training iteration starts by processing a batch of $\mathcal{B}$ FJSP instances, with a new batch resampled every $l_{\text{batch}}$ iterations. For every instance in the batch, the environment initializes a state $s_0$ at time $t=0$, along with initialized prompt about the state, and embedding $\mathbf{z}_0^{\text{LLM}}$. Then, GNN embedding $\mathbf{z}_0^{\text{GNN}}$ is generated and fused together with $\mathbf{z}_0^{\text{LLM}}$ as $\mathbf{h}_0$ as the input of policy network for action selection. At time $t$, the policy samples an action $a_t \sim \pi_\theta(\cdot \mid s_t)$, representing the assignment of a specific operation to a machine. After executing the action, \textsc{Luca} observes the updated state and reward. If the operation's impact exceeds pre-defined thresholds, a hint is appended to the operation’s prompt, which is re-encoded by the LLM to produce a new embedding $\mathbf{z}_{t+1}^{\text{LLM}}$ and eventually fused embedding $\mathbf{h}_{t+1}$ for state $s_{t+1}$. The actor and critic networks are updated before the next batch. The model is evaluated on a held-out replay buffer $\mathcal{I}_{\text{val}}$ and the validation rewards either confirm the policy update or trigger policy rollback. Once trained, the policy $\pi_\theta$ can be deployed for real-time scheduling.

\section{Experimental Study}\label{sec:experiment}
To demonstrate \textsc{Luca}'s performance in solving carbon-aware FJSP, we conduct a comprehensive experimental study in this section. 

\subsection{Experimental Setup}
First, we introduce the experiment setup, which includes datasets, \textsc{Luca} model configurations, and experiment environment. 

\subsubsection{Datasets}
In this study, we use both synthetic and publicly available datasets. First, we follow the standard practice for FJSP and generate a number of instances, each corresponds to a specific carbon-aware FJSP with 10 jobs and 5 machines, where $10 \times 5$ indicates the instance size. Note that \textsc{Luca} has a flexible design and is not restricted to specific sizes of instances. For each operation-machine assignment, the operation processing time is randomly generated. Besides time, each machine is associated with a carbon emission rate. Different machines often have different emission rates. Even for two machines of the same model, the rates may not be the same with different energy sources, historical usage patterns, and ages. Specifically, we randomly choose a rate between $e^{\min}$ and $e^{\max}$, where we fix the former to 1 for consistency and the latter ranges from 2 to 16 to reflect different levels of maximum rate differences. Finally, we organize the instances into distinct sets of training, validation, and testing datasets. 

We further consider two public datasets for comprehensive performance evaluation. One dataset is the well-known \textsc{Mk} with 10 instances \cite{brandimarte1993routing} and the other is \textsc{La} from \cite{behnke2012test}. \textsc{La} consists of three groups, including \textsc{rdata}, \textsc{edata}, and \textsc{vdata}, each has 40 instances. The sizes of the instances vary from $10\times 4$ to $30\times 15$. Emission rate is not available in these datasets and we follow the same scheme as described above to assign an emission rate to each machine. Public datasets do not have many instances and we use synthetic instances only for training.

\subsubsection{Model and Training Configuration} 
At each time slot, \textsc{Luca} adopts LLM and GNN to generate embeddings. For LLM, we use \texttt{all-MiniLM-L6-v2}, which is a pre-trained Sentence Transformer model. It is publicly available in Hugging Face \cite{reimers-2019-sentence-bert}. Given any text prompt, it produces an LLM embedding of a fixed length of 128. For GNN, we configure 2 message passing layers to the graph representation of an instance. The features from the graph is projected into a shared latent space of dimension 8. Integrating both LLM and GNN embeddings, we have the fused embedding which represents the state of the time slot. The state is passed through a neural network with a hidden layer of dimension 128, followed by two separate output heads, one for actor network and the other for critic network. For both networks, the output dimensions are 64, for policy parameters $\theta$ and value parameters $\phi$, respectively.

We configure 1,000 iterations for training, during which, we configure the batch size to 20, i.e., $|\mathcal{B}| = 20$. The initial batch consists of the instances randomly selected and we refresh the batch, e.g., with different instances, for every 20 iterations. For validation, we allocate 100 instances to the held-out replay buffer. For policy training, the PPO loss function consists of three components, including the clipped policy loss with a clip ratio of 0.2, a value loss, and an entropy regularization term. The loss coefficients of the components are set to 1.0, 0.5, and 0.01, respectively. The network is optimized using the \texttt{adam} optimizer with a learning rate of $2 \times 10^{-4}$. Overall, \textsc{Luca} is optimized during the training and validation and converges to a near-optimal scheduling policy. 

All our experiments were conducted on a workstation with an Intel Core Ultra 7 265KF CPU with 20 cores, NVIDIA GeForce RTX 4090 GPU with 24 GB memory, and Ubuntu 24.04.2 LTS 64-bit operating system. 

\subsection{Comparison Study}
We present comparison study in this part where the performance of \textsc{Luca} in terms of both makespan and emission is compared with the performance achieved by other scheduling algorithms. 

\subsubsection{Comparison Algorithms}
We aim to compare \textsc{Luca} with different algorithms and the SOTA based on our knowledge is \textsc{Drl} proposed in \cite{song2022flexible}. However, \textsc{Drl} is not carbon-aware and it is trained for minimizing makespan only. Thus, we implement a variant called \textsc{Drl-c} where \textsc{-c} indicates carbon-awareness. We maintain \textsc{Drl}'s superior solution architecture and settings unchanged for \textsc{Drl-c} except that we replace the makespan-based reward with the same reward function used in \textsc{Luca}, where both makespan and emission are incorporated. Based on Eq. (\ref{eq:reward}), $\lambda$ is the control parameter of the reward function; we set it to 0.5 to assign equal importance to both makespan and emission for model training. We aim to demonstrate that \textsc{Luca} can outperform the carbon-aware SOTA in terms of both makespan and emission. 

To ensure the fairness of comparison, we consider the original version of \textsc{Drl} as well and we implement a variant of \textsc{Luca} which minimizes makespan only, as \textsc{Drl} does, and the variant is denoted as \textsc{Luca-m}. When makespan is the only objective, there are many available algorithms, though most are neither carbon-aware nor optimal for makespan. We choose several of such algorithms such as \textsc{Mor} and \textsc{Spt} in this study. Furthermore, we consider an unrealistic solution based on OR-Tools, which produces schedules with huge computational overhead and impractical time usage, e.g., hours, though its scheduling quality is competitive and its schedules are often treated as the optimal. We denote the solution as \textsc{Offline}. In the following parts, we present experiment results and discussion and let us first compare \textsc{Luca} against the SOTA variant \textsc{Drl-c}.

\begin{table}[h]
\centering
\caption{Performance comparison between \textsc{Luca} and \textsc{Drl-c} on the synthetic dataset. The table reports the mean, STD, and \textsc{Luca}'s mean improvement for makespan and emission. The best results are underlined and \textsc{Luca} performs the best.}
\label{tab:comp-luca-drl-syn}
\small
\renewcommand{\arraystretch}{1.3}
\setlength{\tabcolsep}{6pt}
\begin{tabular}{c|ccc|ccc}
\hline\hline
\multirow{2}{*}{Method}  
& \multicolumn{3}{c|}{Makespan} 
& \multicolumn{3}{c}{Emission} \\\cline{2-7}
& \multicolumn{1}{c}{mean} & \multicolumn{1}{c}{STD} & \multicolumn{1}{c|}{improv.}
& \multicolumn{1}{c}{mean} & \multicolumn{1}{c}{STD} & \multicolumn{1}{c}{improv.} \\
\hline\hline
\textsc{Drl-c} 
& 121.50 & 6.25 & \multicolumn{1}{c|}{$-$} 
& 1062.10 & \underline{254.38} & \multicolumn{1}{c}{$-$} \\\hline
\textsc{Luca} 
& \underline{116.57} & \underline{6.14} & 4.06\% & \underline{1060.36} & 256.30 & 0.16\% \\
\hline\hline
\end{tabular}
\end{table}

\subsubsection{\textsc{Luca} vs. \textsc{Drl-c}}
We first present the results for synthetic dataset in Table \ref{tab:comp-luca-drl-syn}. The statistical results in terms of mean and standard deviation (STD) are based on 10 independent runs of each algorithm and \textsc{Luca}'s improvement over \textsc{Drl-c} is based on the mean results of makespan or emission. We can see that \textsc{Luca} outperforms \textsc{Drl-c} for both makespan and emission, with a 4.1\% improvement in makespan and nearly the same emission, e.g., 0.2\% difference. Both algorithms are RL-based for real-time scheduling and the key differences are the usage of LLM-based embedding and incorporation of carbon factors. \textsc{Luca}'s consistent advantage highlights the effectiveness of incorporating LLM-based embeddings to capture rich semantic representations of instance execution dynamics. \textsc{Drl}, though has been shown to be highly competitive in makespan minimization, struggles to maintain the competitiveness with its \textsc{Drl-c} variant when carbon emission is considered in addition. \textsc{Luca} is able to optimize both makespan and emission well and the results demonstrate its robust performance in handling complex scheduling scenarios for carbon-aware FJSP.

\setlength{\tabcolsep}{3pt}
\begin{table}[h]
\centering
\caption{Comparison between \textsc{Luca} and \textsc{Drl-c} on public datasets. The table reports mean, STD, and \textsc{Luca}’s improvement in makespan and emission. Best results of different settings are underlined. \textsc{Luca} achieves the best makespan overall with minimized emission.}
\label{tab:carbon_pyblic}
\small
\renewcommand{\arraystretch}{1.3}
\setlength{\tabcolsep}{4pt}
\begin{tabular}{c|rrr|rrr}
\hline\hline
\multirow{2}{*}{Dataset} 
& \multicolumn{3}{c|}{Makespan} 
& \multicolumn{3}{c}{Emission} \\\cline{2-7}
& \multicolumn{1}{c}{\textsc{Drl-c}} & \multicolumn{1}{c}{\textsc{Luca}} & \multicolumn{1}{c|}{improv.} & \multicolumn{1}{c}{\textsc{Drl-c}} & \multicolumn{1}{c}{\textsc{Luca}} & \multicolumn{1}{c}{improv.} \\
\hline\hline
\textsc{Mk}         & 216.70 & \underline{202.50} & 6.56\%  & 1987.60 & \underline{1921.01} & 3.35\% \\\hline
\textsc{La-rdata}  & 1118.25 & \underline{1047.28} & 6.36\%  & \underline{12089.43} & 12140.70 & $-$0.42\% \\\hline
\textsc{La-edata}  & 1244.20 & \underline{1233.88} & 0.83\%  & \underline{12101.33} & 12110.55 & $-$0.08\% \\\hline
\textsc{La-vdata}  & 994.93  & \underline{959.70}  & 3.67\%  & 12143.70 & \underline{12107.21} & 0.03\% \\
\hline\hline
\end{tabular}
\end{table}

We further report the results based on the public datasets in Table \ref{tab:carbon_pyblic}. \textsc{Luca} consistently outperforms \textsc{Drl-c} across all datasets for makespan, with performance improvements from 0.8\% for \textsc{La-edata} to 6.6\% for \textsc{Mk}, on average 4.4\% improvement. \textsc{Luca}'s performance for carbon emission is also competitive, with 3.4\% less emission than \textsc{Drl-c} for \textsc{Mk} and the performance difference is minimal, within 0.5\%, for the rest. These results again highlight \textsc{Luca}’s ability to handle complex scenarios effectively. \textsc{Drl-c}, relatively, cannot handle such complexity as well as \textsc{Luca}. Overall, the results demonstrate that \textsc{Luca} is competitive for carbon-aware FJSP and generalized in different settings.

\begin{table}[h]
\caption{Comparison of \textsc{Luca-m} with other methods on the synthetic dataset. Makespan and approximation to the \textsc{Offline} baseline are reported. The best results are underlined and \textsc{Luca-m} performs the best.}
\label{tab:makespan_results}
\small 
\centering
\renewcommand{\arraystretch}{1.3}
\setlength{\tabcolsep}{3pt}
\begin{tabular}{c|c|c|c|c|c|c|c}
\hline \hline
Metric & \textsc{Offline} & \textsc{Luca-m} & \textsc{Drl} & \textsc{Mor} & \textsc{Spt} & \textsc{Fifo} & \textsc{Mwkr} \\
\hline\hline
Makespan & 96.59 & \underline{110.93} & 112.49 & 116.69 & 129.06 & 119.62 & 115.29 \\\hline
\multicolumn{2}{c|}{approx. to \textsc{Offline}}     & \underline{1.148} & 1.165 & 1.208 & 1.336 & 1.238 & 1.194 \\
\hline\hline
\end{tabular}
\end{table}

\subsubsection{\textsc{Luca-m} vs. Other Algorithms} 
In this part, we investigate \textsc{Luca}'s performance in relatively more specialized settings. We specifically consider \textsc{Luca-m}, \textsc{Luca}'s variant focuses on makespan only, as well as the SOTA \textsc{Drl} and several classical heuristics, including \textsc{Mor}, \textsc{Spt}, \textsc{Fifo} and \textsc{Mwkr}. The experiment results on synthetic datasets are shown in Table \ref{tab:makespan_results}, where each algorithm's approximation ($\geq 1$) to the best-known makespan by \textsc{Offline} is included. Seen from the table, \textsc{Luca-m} approximates \textsc{Offline} well, with 1.148 approximation, i.e., 14.8\% longer makespan than the optimal. \textsc{Drl} is the second best algorithm with 1.165 approximation, around 2\% gap to \textsc{Luca-m}. The heuristics exhibit much larger gaps to the near-optimal yet unrealistic \textsc{Offline}, with approximation ranging from 1.194 for \textsc{Mwkr} to 1.336 for \textsc{Spt}, and the gaps to \textsc{Luca-m} range from 5\% to 19\%. The results show that \textsc{Luca-m} is highly competitive even one of \textsc{Luca}'s unique features, carbon-awareness, is not incorporated. 

\begin{table}[h]
\centering
\caption{Comparison of \textsc{Luca-m} with other methods on public benchmarks. Makespan (ms) and approximation (appr.) to the \textsc{Offline} baseline are reported. Best results are underlined. \textsc{Luca-m} performs best overall, except on \textsc{La-vdata}.} 
\label{tab:public_makespan_results}
\small 
\centering
\renewcommand{\arraystretch}{1.3}
\begin{tabular}{l|rr|rr|rr|rr}
\hline \hline
\multirow{2}{*}{Method} & \multicolumn{2}{c|}{\textsc{Mk}} 
                & \multicolumn{2}{c|}{\textsc{La-rdata}} 
                & \multicolumn{2}{c|}{\textsc{La-edata}} 
                & \multicolumn{2}{c}{\textsc{La-vdata}} \\
\cline{2-9}
 & \multicolumn{1}{c}{ms} & appr.  
 & \multicolumn{1}{c}{ms} & appr.  
 & \multicolumn{1}{c}{ms} & appr.  
 & \multicolumn{1}{c}{ms} & appr.  \\
\hline \hline
\textsc{Offline} & 174.00  & --     
         & 938.38  & --     
         & 1026.70 & --     
         & 924.40  & --  \\
\hline
\textsc{Luca-m} & \underline{198.60}  & \underline{1.141}  
             & \underline{1028.38} & \underline{1.096} 
             & \underline{1172.08} & \underline{1.142} 
             & 957.65  & 1.036 \\
\textsc{Drl}     & 201.00  & 1.155  
        & 1030.83 & 1.099   
        & 1187.48 & 1.157 
        & \underline{955.90}  & \underline{1.034} \\
\textsc{Mor}    & 202.31  & 1.163  
        & 1064.04 & 1.134  
        & 1211.23 & 1.180
        & 969.25  & 1.049 \\
\textsc{Spt}    & 238.80  & 1.372  
        & 1184.80 & 1.263  
        & 1305.35 & 1.271 
        & 1085.46 & 1.174 \\
\textsc{Fifo}    & 206.09  & 1.184  
        & 1082.08 & 1.153  
        & 1255.46 & 1.223 
        & 980.69  & 1.061 \\
\textsc{Mwkr}    & 200.17  & 1.150  
        & 1046.20 & 1.115  
        & 1179.90 & 1.149 
        & 962.01  & 1.041 \\
\hline \hline
\end{tabular}
\end{table}

To further evaluate the performance of \textsc{Luca-m}, \textsc{Drl}, and the heuristics, we test on public datasets and the results are shown in Table \ref{tab:public_makespan_results}. We can see that \textsc{Luca-m} performs the best on most datasets. For \textsc{Mk}, \textsc{Luca-m} approximates the optimal by 1.141, and the approximations are 1.096 and 1.142 for \textsc{La-rdata} and \textsc{La-edata}, respectively. \textsc{Drl} is the best algorithm for \textsc{La-vdata} and \textsc{Luca-m} is only less than 0.2\% behind. The results on both synthetic and public datasets suggest that \textsc{Luca}'s adoption of the RL framework can be justified and LLM provides an enhanced understanding of scheduling tasks with our effective mechanisms, e.g., adding hints about the operations with high makespan impact.

\subsection{Sensitivity Analysis}
In this part, we perform sensitivity analysis for \textsc{Luca} with different parameter settings.

\begin{figure}[]
    \centering
    \includegraphics[width=1\linewidth]{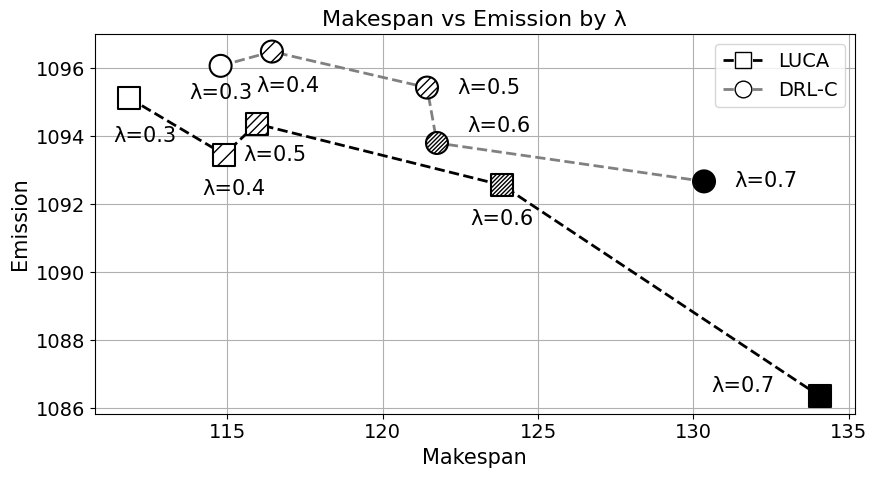}
    \caption{Multi-objective performance comparison between \textsc{Luca} and \textsc{Drl-c} across various values of the balance parameter $\lambda$. \textsc{Luca} yields Pareto-dominant schedules.}
    \label{fig:lambda}
\end{figure}

\subsubsection{Sensitivity to $\lambda$}
\textsc{Luca} optimizes both makespan and emission and the relative importance of the two objectives is controlled by parameter $\lambda$, as shown in Eq. (\ref{eq:reward}). We consider different values of $\lambda$ from 0.3 to 0.7, with 0.5 indicating a balanced importance of the two objectives. The results based on \textsc{Luca} and \textsc{Drl-c} are illustrated in Figure \ref{fig:lambda}, where we connect the points of discrete $\lambda$ values to well visualize the trends of makespan and emission optimization achieved by each algorithm. We can see that increasing the value of $\lambda$ generally results in decreased emissions and increased makespan. This observation aligns with Eq. (\ref{eq:reward}) where a high value of $\lambda$ emphasizes emission, and vice verse. There is a trade-off between makespan and emission for each algorithm. Relatively, \textsc{Luca}'s trade-off is closer to the bottom left, indicating \textsc{Luca}'s advantages in comparison with \textsc{Drl-c}. For example, given small $\lambda \leq 0.5$, makespan is important and \textsc{Luca}'s makespan is consistently lower than \textsc{Drl-c}, up to 4.1\% better for $\lambda=0.5$. Finally, we argue that choosing an appropriate value of $\lambda$ is crucial for balancing the two objectives for real-world applications; values too large or too small may favor one objective at the potentially significant expense of the other.

\begin{table}[]
\centering
\small
\renewcommand{\arraystretch}{1.3}
\setlength{\tabcolsep}{5pt}
\caption{Comparison of \textsc{Luca} and \textsc{Drl-c} under different emission ratios. A high ratio indicates big rates variation. The table shows average makespan and emissions for both methods. The best results are underlined. \textsc{Luca} performs the best.
}
\label{tab:emission_ratio}
\begin{tabular}{c|ccc|rrc}
\hline \hline
Emission
& \multicolumn{3}{c|}{Makespan} 
& \multicolumn{3}{c}{Emission} \\ \cline{2-7}
Ratio & \textsc{Luca} & \textsc{Drl-c} & improv. 
& \multicolumn{1}{c}{\textsc{Luca}} & \multicolumn{1}{c}{\textsc{Drl-c}} & improv. \\
\hline \hline
$1:2$   & \underline{112.90} & 118.93 & 5.07\% & \underline{982.29}  & 983.32  & 0.11\% \\
$1:4$   & \underline{114.22} & 117.33 & 2.65\% & \underline{1260.70} & 1261.41 & 0.06\% \\
$1:8$   & \underline{114.24} & 119.45 & 4.36\% & \underline{1175.81} & 1177.97 & 0.18\% \\
$1:16$  & \underline{115.95} & 121.43 & 4.52\% & \underline{1094.36} & 1095.43 & 0.10\% \\
\hline \hline
\end{tabular}
\end{table}

\subsubsection{Sensitivity to Emission Rate} 
Each machine in carbon-aware FJSP is assigned with an emission rate and the rate variation is controlled by the ratio between minimum and maximum rates. We investigate \textsc{Luca}'s performance robustness with different ratios, from $1:2$ to $1:16$, and show the results in Table \ref{tab:emission_ratio} with $\lambda$ fixed at 0.5. We can see that \textsc{Luca}'s performance remains competitive across different ratios, outperforming \textsc{Drl-c} by 4.2\% on average and up to 5.1\% for makespan and achieving on average 0.11\% lower emissions. Makespan is relatively large with big ratios, which drives \textsc{Luca} to choose machines with low emission rate and thus may lead to certain sacrifice in processing time.  Overall, this sensitivity analysis highlight \textsc{Luca}'s performance robustness under varying carbon intensity conditions.

\balance
\section{Conclusion}
\label{sec:conclusion}
This paper introduced \textsc{Luca}, an LLM augmented graph deep RL framework for carbon-aware FJSP. By fusing structural features captured by GNN with contextual information from an LLM through our in-house prompting scheme, \textsc{Luca} effectively captures scheduling dynamics. \textsc{Luca} is RL-based, where its policy is optimized by a dual-objective reward function for both makespan and carbon emission. Extensive evaluations on synthetic and public datasets confirm the effectiveness and robustness of \textsc{Luca}. It consistently outperforms comparison algorithms in different settings, achieving significant reductions in makespan, e.g., on average 4.1\% and up to 12.2\% better than \textsc{Drl-c} while maintaining or improving emission levels. It outperforms the comparison algorithms for makespan under realistic settings even when one of its competitive features, carbon-awareness, is disabled. These results highlight \textsc{Luca}’s potential for real-time and sustainable scheduling in smart manufacturing system. Besides, \textsc{Luca} has a generic design for incorporating LLM into the RL framework, which also shed lights on scheduling applications in other industry sectors.

Building on the promising results of \textsc{Luca}, several directions remain for future exploration. Real-time feedback from physical manufacturing environments can be incorporated to further enhance the adaptability and robustness of a scheduling policy. The interpretability of the LLM-generated prompts and their influence on decision-making and control can be further investigated to improve transparency and trust in ML-based scheduling. Additionally, integrating other sustainability factors, such as varying electricity prices, could lead to more holistic optimization in smart manufacturing applications.

\begin{acks}
This research is supported by A*STAR under its MTC Individual Research Grants (IRG) (Award M23M6c0113), MTC Programmatic (Award M23L9b0052), SIT’s Ignition Grant (STEM) (Grant ID: IG (S) 2/2023 – 792), and the National Research Foundation Singapore and DSO National Laboratories under the AI Singapore Programme (AISG Award No: AISG2-GC-2023-006).
\end{acks}

\bibliographystyle{ACM-Reference-Format}
\bibliography{reference}

@article{zhao2025diffusion,
  title={Diffusion-Modeled Reinforcement Learning for Carbon and Risk-Aware Microgrid Optimization},
  author={Zhao, Yunyi and Zhang, Wei and Xiang, Cheng and Du, Hongyang and Niyato, Dusit and Gao, Shuhua},
  journal={arXiv preprint arXiv:2507.16867},
  year={2025}
}

@article{zhang2024role,
  title={The role of generative artificial intelligence in internet of electric vehicles},
  author={Zhang, Hanwen and Niyato, Dusit and Zhang, Wei and Zhao, Changyuan and Du, Hongyang and Jamalipour, Abbas and Sun, Sumei and Pei, Yiyang},
  journal={IEEE Internet of Things Journal},
  year={2024},
  publisher={IEEE}
}

@article{wang2022cea,
  title={CEA-FJSP: Carbon emission-aware flexible job-shop scheduling based on deep reinforcement learning},
  author={Wang, Shiyong and Li, Jiaxian and Tang, Hao and Wang, Juan},
  journal={Frontiers in Environmental Science},
  volume={10},
  pages={1059451},
  year={2022},
  publisher={Frontiers Media SA}
}

@article{zhang2024deep,
  title={Deep reinforcement learning-based memetic algorithm for energy-aware flexible job shop scheduling with multi-AGV},
  author={Zhang, Fayong and Li, Rui and Gong, Wenyin},
  journal={Computers \& Industrial Engineering},
  volume={189},
  pages={109917},
  year={2024},
  publisher={Elsevier}
}

@article{zhao2024actor,
  title={An actor-critic framework based on deep reinforcement learning for addressing flexible job shop scheduling problems},
  author={Zhao, Cong and Deng, Na},
  journal={Math. Biosci. Eng},
  volume={21},
  number={1},
  pages={1445--1471},
  year={2024}
}

@article{brandimarte1993routing,
  title={Routing and scheduling in a flexible job shop by tabu search},
  author={Brandimarte, Paolo},
  journal={Annals of Operations research},
  volume={41},
  number={3},
  pages={157--183},
  year={1993},
  publisher={Springer}
}

@article{behnke2012test,
  title={Test instances for the flexible job shop scheduling problem with work centers},
  author={Behnke, Dennis and Geiger, Martin Josef},
  year={2012},
  publisher={Institut f{\"u}r betriebliche Logistik und Organisation}
}

@misc{worldbank_sg_manu,
  author= {{The World Bank}},
  title= {{Manufacturing, value added (\% of GDP) – Singapore}},
  year= {2023},
  howpublished ={\url{https://data.worldbank.org.}},
  note= {Accessed: 2025-06-30}
}

@article{chen2013flexible,
  title={A flexible dispatching rule for minimizing tardiness in job shop scheduling},
  author={Chen, Binchao and Matis, Timothy I},
  journal={International Journal of Production Economics},
  volume={141},
  number={1},
  pages={360--365},
  year={2013},
  publisher={Elsevier}
}

@article{montazeri1990analysis,
  title={Analysis of scheduling rules for an FMS},
  author={Montazeri, Mrn and Van Wassenhove, LN},
  journal={The International Journal of Production Research},
  volume={28},
  number={4},
  pages={785--802},
  year={1990},
  publisher={Taylor \& Francis}
}

@article{song2022flexible,
  title={Flexible job-shop scheduling via graph neural network and deep reinforcement learning},
  author={Song, Wen and Chen, Xinyang and Li, Qiqiang and Cao, Zhiguang},
  journal={IEEE Transactions on Industrial Informatics},
  volume={19},
  number={2},
  pages={1600--1610},
  year={2022},
  publisher={IEEE}
}

@article{piroozfard2018minimizing,
  title={Minimizing total carbon footprint and total late work criterion in flexible job shop scheduling by using an improved multi-objective genetic algorithm},
  author={Piroozfard, Hamed and Wong, Kuan Yew and Wong, Wai Peng},
  journal={Resources, Conservation and Recycling},
  volume={128},
  pages={267--283},
  year={2018},
  publisher={Elsevier}
}

@article{liu2017hybrid,
  title={A hybrid fruit fly algorithm for solving flexible job-shop scheduling to reduce manufacturing carbon footprint},
  author={Liu, Qiong and Zhan, Mengmeng and Chekem, Freddy O and Shao, Xinyu and Ying, Baosheng and Sutherland, John W},
  journal={Journal of Cleaner Production},
  volume={168},
  pages={668--678},
  year={2017},
  publisher={Elsevier}
}

@misc{source_downtime_loss,
  author       = {FourJaw},
  title        = {The Cost of Downtime in Manufacturing},
  year         = {2024},
  howpublished = {\url{https://fourjaw.com/blog/the-cost-of-downtime-in-manufacturing}},
  note         = {Accessed: 2025-06-30}
}

@article{abgaryan2024llms,
  title={Llms can schedule},
  author={Abgaryan, Henrik and Harutyunyan, Ararat and Cazenave, Tristan},
  journal={arXiv preprint arXiv:2408.06993},

  year={2024}
}

@inproceedings{reimers-2019-sentence-bert,
  title = "Sentence-BERT: Sentence Embeddings using Siamese BERT-Networks",
  author = "Reimers, Nils and Gurevych, Iryna",
  booktitle = "Proceedings of the 2019 Conference on Empirical Methods in Natural Language Processing",
  month = "11",
  year = "2019",
  publisher = "Association for Computational Linguistics",
}

\end{document}